\documentclass[runningheads, envcountsame, a4paper]{llncs}
\usepackage[pdftex]{graphicx}
\usepackage{epstopdf}
\usepackage[hyphens]{url}
\usepackage{microtype}

\usepackage{amssymb}
\usepackage{amsmath}
\usepackage{bbm}
\usepackage{multicol}
\usepackage{adjustbox}
\usepackage{mathtools}
\usepackage{multirow}
\usepackage{appendix}
\usepackage[caption=false]{subfig}
\usepackage{array}
\usepackage{bm}
\usepackage[misc]{ifsym}
\usepackage{xcolor}
\usepackage{placeins}

\begin{document}
\title{Inter-domain Multi-relational Link Prediction}
\toctitle{Inter-domain Multi-relational Link Prediction}
%
%
\author{Luu Huu Phuc\inst{1}(\Letter)\and Koh Takeuchi\inst{1}\and Seiji Okajima\inst{2}\and Arseny Tolmachev\inst{2}\and Tomoyoshi Takebayashi\inst{2}\and Koji Maruhashi\inst{2}\and Hisashi Kashima\inst{1}}
\authorrunning{L.H. Phuc et al.}
\tocauthor{Luu~Huu~Phuc, Koh~Takeuchi, Seiji~Okajima, Arseny~Tolmachev, Tomoyoshi~Takebayashi, Koji~Maruhashi, Hisashi~Kashima}
\institute{Kyoto University, Japan \\
\email{$\{$phuc, koh, kashima$\}$@ml.ist.i.kyoto-u.ac.jp}\\\and
Fujitsu Research, Fujitsu Ltd., Japan \\
\email{$\{$okajima.seiji, t.arseny, takebayashi.tom, maruhashi.koji$\}$@fujitsu.com}\\
}
\maketitle              
\setcounter{footnote}{0}
\begin{abstract}
Multi-relational graph is a ubiquitous and important data structure, allowing flexible representation of multiple types of interactions and relations between entities. Similar to other graph-structured data, link prediction is one of the most important tasks on multi-relational graphs and is often used for knowledge completion. When related graphs coexist, it is of great benefit to build a larger graph via integrating the smaller ones. The integration requires predicting hidden relational connections between entities belonged to different graphs (inter-domain link prediction). However, this poses a real challenge to existing methods that are exclusively designed for link prediction between entities of the same graph only (intra-domain link prediction). In this study, we propose a new approach to tackle the inter-domain link prediction problem by softly aligning the entity distributions between different domains with optimal transport and maximum mean discrepancy regularizers. Experiments on real-world datasets show that optimal transport regularizer is beneficial and considerably improves the performance of baseline methods.

\end{abstract}
\section{Introduction}

Multi-relational data represents knowledge about the world and provides a graph-like structure of this knowledge. It is defined by a set of entities and a set of predicates between these entities. The entities can be objects, events, or abstract concepts while the predicates represent relationships involving two entities. 
A multi-relational data contains a set of facts represented as triplets $(e_h, r, e_t)$ denoting the existence of a predicate $r$ from subject entity $e_h$ to object entity $e_t$. In a sense, multi-relational data can also be seen as a directed graph with multiple types of links (multi-relational graph).

A multi-relational graph is often very sparse with only a small subset of true facts being observed. Link prediction aims to complete a multi-relational graph by predicting new hidden true facts based on the existing ones. Many existing methods follow an embedding-based approach which has been proved to be effective for multi-relational graph completion. These methods all aim to find reasonable embedding presentations for each entity (node) and each predicate (type of link). In order to predict if a fact $(e_h, r, e_t)$ holds true, they use a scoring function whose inputs are embeddings of the entities $e_h, e_t$ and the predicate $r$ to compute a prediction score. Some of the most prominent methods in that direction are TransE~\cite{transE}, RESCAL~\cite{rescal}, DisMult~\cite{disMult}, and NTN~\cite{NTN}, to name a few.

TransE~\cite{transE} model is inspired by the intuition from Word2Vec~\cite{word2vec1,word2vec2} that many predicates represent linear translations between entities in the latent embedding space, e.g. $\mathbf{a}_\text{Japan} - \mathbf{a}_\text{Tokyo} \approx \mathbf{a}_\text{Germany} - \mathbf{a}_\text{Berlin} \approx \mathbf{a}_\text{is\_capital\_of}$. Therefore, TransE tries to learn low-dimensional and dense embedding vectors so that $\mathbf{a}_h + \mathbf{a}_r \approx \mathbf{a}_t$ for a true fact $(e_h, r, e_t)$. Its scoring function is defined accordingly via $\lVert \mathbf{a}_h + \mathbf{a}_r - \mathbf{a}_t \rVert_2$.
RESCAL~\cite{rescal} is a tensor factorization-based method. It converts a multi-relational graph data into a $3$-D tensor whose first two modes indicate the entities and the third mode indicates the predicates. A low-rank decomposition technique is employed by RESCAL to compute embedding vectors $\mathbf{a}$ of the entities and embedding matrices $\mathbf{R}$ of the predicates. Its scoring function is the bilinear product $\mathbf{a}_h^\top \mathbf{R}_r \mathbf{a}_t$.
DistMult~\cite{disMult} is also a bilinear model and is based on RESCAL where each predicate is only represented by a diagonal matrix rather than a full matrix.
The neural tensor network (NTN) model~\cite{NTN} generalizes RESCAL's approach by combining traditional MLPs and bilinear operators to represent each relational fact.

Despite achieving state of the art for link prediction tasks, existing methods are exclusively designed and limited to intra-domain link prediction. They only consider the case in which both entities belong to the same relational graph (intra-domain). When the needs for predicting hidden facts between entities of different but related graphs (inter-domain) arise, unfortunately, the existing methods are inapplicable. One of such examples is when it is necessary to build a large relational graph by integrating several existing smaller graphs whose entity sets are related. This study proposes to tackle the inter-domain link prediction problem by learning suitable latent embeddings that minimize dissimilarity between the domains' entity distributions. 

Two popular divergences, namely optimal transport's Wasserstein distance (WD) and the maximum mean discrepancy (MMD), are investigated. 
Given two probability distributions, optimal transport computes an optimal transport plan that gives the minimum total transport cost to relocate masses between the distributions. The minimum total transport cost is often known under the name of Wasserstein distance.
In a sense, the computed optimal transport plan and the corresponding Wasserstein distance
provide a reasonable alignment and quantity for measuring the dissimilarity between the supports/domains of the two distributions. 
Minimizing Wasserstein distance has been proved to be effective in enforcing the alignment of corresponding entities across different domains and is successfully applied in graph matching~\cite{gw-learning}, cross-domain alignment~\cite{got}, and multiple-graph link prediction problems~\cite{self-cite}. 
As another popular statistical divergence between distributions, MMD computes the dissimilarity by comparing the kernel mean embeddings of two distributions in a reproducing kernel Hilbert space (RKHS). It has been widely applied in two-sample tests for differentiating distributions~\cite{MMD-original1,MMD-original2} and distribution matching in domain adaptation tasks~\cite{DMM}, to name a few.

The proposed method considers a setting of two multi-relational graphs whose entities are assumed to follow the same underlying distribution. 
For example, the multi-relational graphs can be about relationships among users/items in different e-commerce flatforms of the same country. 
They could also be knowledge graphs of semantic relationships between general concepts that are built from different common-knowledge sources, e.g. Freebase and DBpedia.
In both examples, it is safe to assume that the entity sets are distributionally identical.
This assumption is fundamental for the regularizers to be effective in connecting the entity distributions of the two graphs.

\section{Preliminary}

This section briefly introduces the components that are employed in the proposed method.

\subsection{RESCAL}

RESCAL~\cite{rescal} formulates a multi-relational data as a three-way tensor $\mathbf{X} \in \mathbb{R}^{n \times n \times m}$, where $n$ is the number of entities and $m$ is the number of predicates. $\mathbf{X}_{i,j,k} = 1$ if the fact $(e_i, r_k, e_j)$ exists and $\mathbf{X}_{i,j,k} = 0$ otherwise. 
In order to find proper latent embeddings for the entities and the predicates, RESCAL performs a rank-$d$ factorization where each slice along the third mode $\mathcal{X}_k = \mathbf{X}_{\cdot,\cdot,k}$ is factorized as
\[
\mathcal{X}_k \approx \mathbf{A} \mathbf{R}_k \mathbf{A}^\top ,~~~ \text{for}~k = 1,...,m.
\]
Here, $\mathbf{A} = [\mathbf{a}_1, ..., \mathbf{a}_n]^\top \in \mathbb{R}^{n \times d}$ contains the latent embedding vectors of the entities and $\mathbf{R}_k \in \mathbb{R}^{d \times d}$ is an asymmetric matrix that represents the interactions between entities in the $k$-th predicate.

Originally, it is proposed to learn $\mathbf{A}$ and $\mathbf{R}_k$ with the regularized squared loss function
\[
\min_{\mathbf{A}, \mathbf{R}_k} g(\mathbf{A}, \mathbf{R}_k) + \text{reg}(\mathbf{A}, \mathbf{R}_k), 
\]
where
\[
g(\mathbf{A}, \mathbf{R}_k) = \frac{1}{2} \left(\sum_{k} \lVert \mathcal{X}_k - \mathbf{A} \mathbf{R}_k \mathbf{A}^\top \rVert_{F}^2 \right) 
\]
and $\text{reg}$ is the following regularization term
\[
\text{reg}(\mathbf{A}, \mathbf{R}_k) = \frac{1}{2} \mu \left( \lVert \mathbf{A} \rVert_F^2 + \sum_k \lVert \mathbf{R}_k \rVert_F^2 \right).
\]
$\mu > 0$ is a hyperparameter.

It is later proposed by the authors of RESCAL to learn the embeddings with pairwise loss training~\cite{nickel_review}, i.e. using the following margin-based ranking loss function
\begin{equation}
\min_{\mathbf{A}, \mathbf{R}_k} L(\mathbf{A}, \mathbf{R}_k) = \sum_{(e_i, r_k, e_j) \in \mathcal{D}^+} \sum_{(e_l, r_h, e_t) \in \mathcal{D}^-} \mathcal{L}(f_{ijk}, f_{lth}) + \text{reg}(\mathbf{A}, \mathbf{R}_k), 
\label{eq:rescal}
\end{equation}
where $\mathcal{D}^+$ and $\mathcal{D}^-$ are the sets of all positive triplets (true facts) and all negative triplets (false facts), respectively.
$f_{ijk}$ denotes the score of $(e_i, r_k, e_j)$, $f_{ijk} = \mathbf{a}_i^\top \mathbf{R}_k \mathbf{a}_j$ and $\mathcal{L}$ is the ranking function 
\[
\mathcal{L}(f^+, f^-) = \max(1 + f^- - f^+, 0).
\]
The negative triplet set $\mathcal{D}^-$ is often generated by corrupting positive triplets, i.e. replacing one of the two entities in a positive triplet $(e_i, r_k, e_j)$ with a randomly sampled entity. 

The pairwise loss training aims to learn $\mathbf{A}$ and $\mathbf{R}_k$ so that the score $f^+$ of a positive triplet is higher than the score $f^-$ of a negative triplet. Moreover, the margin-based ranking function is more flexible and easier to optimize with \textit{stochastic gradient descent} (SGD) than the original squared loss function. In the proposed method, the pairwise loss training is adopted.

\subsection{Optimal Transport}

Given two probability vectors $\boldsymbol{\pi}_1 \in \mathbb{R}_{+}^{n_1}$ and $\boldsymbol{\pi}_2 \in \mathbb{R}_{+}^{n_2}$ that satisfy $\boldsymbol{\pi}_1^\top \mathbbm{1}_{n_1} = \boldsymbol{\pi}_2^\top \mathbbm{1}_{n_2} = 1$, a matrix $\mathbf{P} \in \mathbb{R}_{+}^{n_1 \times n_2}$ is called a transport plan between $\boldsymbol{\pi}_1$ and $\boldsymbol{\pi}_2$ if $\mathbf{P} \mathbbm{1}_{n_2} = \boldsymbol{\pi}_1$ and $\mathbf{P}^\top \mathbbm{1}_{n_1} = \boldsymbol{\pi}_2$. Here, $\mathbbm{1}_n$ indicates a $n$-dimensional vector of ones. 
Let's denote the supports of $\boldsymbol{\pi}_1$ and $\boldsymbol{\pi}_2$ as $\mathbf{A}^1 = [\mathbf{a}^1_1, ..., \mathbf{a}^1_{n_1}]^\top \in \mathbb{R}^{n_1 \times d}$ and $\mathbf{A}^2 = [\mathbf{a}^2_1, ..., \mathbf{a}^2_{n_2}]^\top \in \mathbb{R}^{n_2 \times d}$, respectively. 
A transport cost $\mathbf{C} \in \mathbb{R}_{+}^{n_1 \times n_2}$ can be defined as 
\[
C_{ij} = \lVert \mathbf{a}_i^{1} - \mathbf{a}_j^{2} \rVert_2^2.
\]
Given a transport matrix $C$, the transport cost of a transport plan $\textbf{P}$ is computed by
\[
\langle \mathbf{P}, \mathbf{C} \rangle = \sum_{i,j} P_{ij} C_{ij} .
\]
A transport plan $\mathbf{P}^*$ that gives the minimum transport cost, $\mathbf{P}^* = \arg \min_{\mathbf{P}}\langle \mathbf{P}, \mathbf{C} \rangle$, is called an optimal transport plan and the corresponding minimum cost is called the Wasserstein distance. The optimal transport plan $\mathbf{P}^*$ gives a reasonable ``soft" matching between the two distributions $(\boldsymbol{\pi}_1, \mathbf{A}_1)$ and $(\boldsymbol{\pi}_2, \mathbf{A}_2)$ while the Wasserstein distance provides a measurement of how far the two distributions are from each other.

In the scope of multi-relational graphs, $\boldsymbol{\pi}_1$ and $\boldsymbol{\pi}_2$ are predefined over the sets of entities, normally being set to be uniform and the supports $\mathbf{A}_1$ and $\mathbf{A}_2$ can be seen as embeddings of the entities.

The computational complexity of computing the optimal transport plan and Wasserstein distance is often prohibitive.
An efficient approach to compute an approximation has been proposed by Cuturi et al.~\cite{sinkhorn}. Instead of the exact optimal transport $\mathbf{P}^*$, they compute an entropic-regularized transport plan $\mathbf{P}^{\lambda}$ via minimizing a cost $M$ as follows,
\begin{equation}
\mathbf{P}^{\lambda} = \arg \min_{\mathbf{P}} M(\mathbf{P}) = \langle \mathbf{P} , \mathbf{C} \rangle + \frac{1}{\lambda} \sum_{i,j} P_{ij} \log P_{ij},
\label{eq:ot}
\end{equation}
where $\lambda > 0$ is a hyperparameter controlling the effect of the negative entropy of matrix $\mathbf{P}$. With large enough $\lambda$, emperically when $\lambda > 50$, $\mathbf{P}^*$ and the Wasserstein distance can be accurately approximated by $\mathbf{P}^{\lambda}$ and $M(\mathbf{P}^{\lambda})$. 

$\mathbf{P}^{\lambda}$ has a unique solution of the following form
\[
\mathbf{P}^{\lambda} = \textbf{diag}(\mathbf{u}) \mathbf{K} \textbf{diag}(\mathbf{v}),
\]
where $\textbf{diag(\textbf{u})}$ indicates a diagonal matrix whose diagonal elements are elements of $\textbf{u}$. The matrix $\mathbf{K} = e^{-\lambda \mathbf{C}}$ is the element-wise exponential of $-\lambda \mathbf{C}$. Vectors $\mathbf{u}$ and $\mathbf{v}$ can be initialized randomly and updated via Sinkhorn iteration
\[
(\mathbf{u}, \mathbf{v}) \leftarrow \left( \frac{\boldsymbol{\pi}_1}{\mathbf{K} \mathbf{v}}, \frac{\boldsymbol{\pi}_2}{\mathbf{K}^\top \mathbf{u}} \right).
\]

\subsection{Maximum Mean Discrepancy}

Maximum Mean Discrepancy (MMD) is originally introduced as a non-parametric statistic to test if two distributions are different~\cite{MMD-original1,MMD-original2}. It is defined as the difference between mean function values on samples generated from the distributions. If MMD is large, the two distributions are likely to be distinct. On the other hand, if MMD is small, the two distributions can be seen to be similar. Formally, let $\boldsymbol{\pi}_1$ and $\boldsymbol{\pi}_2$ be two distributions whose the supports are subsets of $\mathbb{R}^d$, and $\mathcal{F}$ be a class of functions $f: \mathbb{R}^d \rightarrow \mathbb{R}$. Usually, $\mathcal{F}$ is selected to be the unit ball in a universal RKHS $\mathcal{H}$. Then MMD is defined as 
\[
M(\mathcal{F}, \boldsymbol{\pi}_1, \boldsymbol{\pi}_2) = \sup_{f \in \mathcal{F}} \left( \mathbb{E}_{x \sim \boldsymbol{\pi}_1}[f(x)] - \mathbb{E}_{y \sim \boldsymbol{\pi}_2}[f(y)] \right).
\]

From sample sets $\mathbf{A}^1 = \{\mathbf{a}_1^1, ..., \mathbf{a}_{n_1}^1\}$ and $\mathbf{A}^2 = \{\mathbf{a}_1^2, ..., \mathbf{a}_{n_2}^2\}$, $\mathbf{a}^t_i \in \mathbb{R}^d$, sampled from the two distributions, MMD can be unbiasedly approximated using Gaussian kernels $k(\cdot, \cdot)$ as follows~\cite{computational-OT,MMD-original1}. 
\begin{equation}
\begin{split}
    M(\mathbf{A}^1, \mathbf{A}^2) = &
    \frac{1}{n_1 (n_1 - 1)} \sum_{i, i'} k(\mathbf{a}^1_i, \mathbf{a}^1_{i'}) +
    \frac{1}{n_2 (n_2 - 1)} \sum_{j, j'} k(\mathbf{a}^2_j, \mathbf{a}^2_{j'}) \\ 
    & - \frac{2}{n_1 n_2} \sum_{i,j} k(\mathbf{a}^1_i, \mathbf{a}^2_j)
\end{split}
\label{eq:mmd}
\end{equation}
When $\mathbf{A}^1$ and $\mathbf{A}^2$ are the embeddings of entities in two domains, MMD represents a dissimilarity between the domains' entity distributions.

\section{Problem Setting and Proposed Method}

\subsection{Problem Setting}

The formal problem setting considered in this study is stated as follows.
Given two multi-relational graphs $G^1$ and $G^2$, each graph $G^t$ is defined with a set of entities (nodes) $\mathcal{E}^t = \{e^t_1, ..., e^t_{n_t}\}$, a set of predicates (types of links) $\mathcal{R}^t = \{r^t_1, ..., r^t_{m_t}\}$, and a set of true facts (observed links) $\mathcal{T}^t = \{ (e^t_i, r^t_k, e^t_j) \}$ for $t \in \{1, 2\}$. For simplicity, this study only considers the case where the two graphs share the same set of predicates, i.e. $\mathcal{R}^1 \equiv \mathcal{R}^2 \equiv \mathcal{R}$.
The goal is to predict if an inter-domain fact $(e^1_i, r_k, e^2_j)$ or $(e^2_i, r_k, e^1_j)$ holds true or not.

The entity embeddings of the two graphs are assumed to follow the same distribution, i.e. there exists a distribution $\boldsymbol{\pi}$ such that $\mathbf{a}^t_i \sim \boldsymbol{\pi}$ for embedding $\mathbf{a}^t_i$ of entity $e^t_i \in \mathcal{E}^t$.
In the experiments, the entity sets $\mathcal{E}^1$ and $\mathcal{E}^2$ are controlled so that they are completely disjoint or partially overlapped with only a small amount of common entities. The common entities are known in overlapping settings.

\subsection{Proposed objective function}

The proposed method's objective function consists of two components. The first component is for learning embedding representations of the entities and the predicates of each multi-relational graph, which is based on an existing tensor-factorization method. RESCAL~\cite{rescal} is specifically chosen in the proposed method. The second component is a regularization term for enforcing the entity embedding distributions of the two graphs to become similar.

For each graph $G^t$, lets denote the entity embeddings as $\mathbf{A}^t = [\mathbf{a}^t_1, ..., \mathbf{a}^t_{n_t}]^\top \in \mathbb{R}^{n_t \times d}$, where $d$ is the embedding dimension. 
If the entity sets $\mathcal{E}^1$ and $\mathcal{E}^2$ overlap, the embeddings of common entities are set to be identical in both domains, i.e. $\mathbf{A}^t = [\mathbf{A}^{'t}, \mathbf{A}_c]^\top$ where $\mathbf{A}_c \in \mathbb{R}^{d \times |\mathcal{E}^1 \cap \mathcal{E}^2|}$ is the embeddings of common entities.
The embedding of predicate $r_k \in \mathcal{R}$ is denoted as $\mathbf{R}_k \in \mathbb{R}^{d \times d}$ for $k\in\{1,...,m\}$. 
The objective function of the proposed method is given as
\begin{equation}
F(\mathbf{A}^1, \mathbf{A}^2, \mathbf{R}_k, [\mathbf{P}]) = 
L(\mathbf{A}^1, \mathbf{R}_k) + L(\mathbf{A}^2, \mathbf{R}_k) + \alpha M(\mathbf{A}^1, \mathbf{A}^2, [\mathbf{P}]).
\label{eq:objective}
\end{equation}
In \eqref{eq:objective}, the first two terms $L(\mathbf{A}^t, \mathbf{R}_k)$ are the loss functions of RESCAL and are defined as in \eqref{eq:rescal}. 
The third term $M(\mathbf{A}^1, \mathbf{A}^2, [\mathbf{P}])$ is the entropic-regularized Wasserstein distance (WD) or the MMD discrepancy between the entity distributions of the two graphs. 
In the case of WD regularizer, $M = M(\mathbf{A}^1, \mathbf{A}^2, \mathbf{P})$ is defined as in \eqref{eq:ot} with $\mathbf{P} \in \mathbb{R}_+^{n_1 \times n_2}$.
In the case of MMD regularizer, $M = M(\mathbf{A}^1, \mathbf{A}^2)$ is defined as in \eqref{eq:mmd}.

Via $L(\mathbf{A}^t, \mathbf{R}_k)$, the underlying embedding distribution over each entity set $\mathcal{E}^t$ is learned and characterized into $\mathbf{A}^t$, while $M(\mathbf{A}^1, \mathbf{A}^2, [\mathbf{P}])$ helps to drive these two distributions to become similar.
Through the objective function $F$, similar entities of $G^1$ and $G^2$ are expected to 
lie close to each other on the latent embedding space, 
which encourages similar entities to involve in similar relations/links. Specifically, if $e^1_i \in \mathcal{E}^1$ and $e^2_i \in \mathcal{E}^2$ have similar embeddings $\mathbf{a}^1_i$ and $\mathbf{a}^2_i$, the inter-domain fact $(e^1_i, r_k, e^2_j)$ is likely to exist if the intra-domain fact $(e^2_i, r_k, e^2_j)$ exists thanks to their similar scores ${\mathbf{a}^1_i}^\top \mathbf{R}_k \mathbf{a}^2_j \approx {\mathbf{a}^2_i}^\top \mathbf{R}_k \mathbf{a}^2_j$.

The objective function $F(\mathbf{A}^1, \mathbf{A}^2, \mathbf{R}_k)$ (MMD regularizer) is directly optimized with SGD. On the other hand, $F(\mathbf{A}^1, \mathbf{A}^2, \mathbf{R}_k, \mathbf{P})$ (WD regularizer) is minimized iteratively. In each epoch, the transport plan $\mathbf{P}$ is fixed and the embedding vectors $\mathbf{A}^1$ and $\mathbf{A}^2$ are updated with SGD. At the end of each epoch, $\mathbf{A}^1$ and $\mathbf{A}^2$ are fixed and the plan $\mathbf{P}$ is sequentially updated via Sinkhorn algorithm~\cite{sinkhorn}.

\section{Experiments}

\subsection{Datasets}

The datasets used in the experiments are created from four popular knowledge graph datasets, namely FB15k-237~\cite{fb15k}, WN18RR~\cite{wn18rr}, DBbook2014, and ML1M~\cite{dbbook-ml1m}. 
The FB15k-237 dataset contains $272k$ facts about general knowledge. It has $14k$ entities and $237$ predicates.
The WN18RR dataset consists of $86k$ facts about $11$ lexical relations between $40k$ word senses. 
The other two datasets represent interactions among users and items in e-commerce.
The ML1M (MovieLens-1M) dataset composes of $434k$ facts with $14k$ users/items and $20$ relations, while the DBbook2014 has $334k$ facts with $13k$ users/items and $13$ relations.
To create $G^1$ and $G^2$ for each dataset, two smaller sub-graphs of around $2k$ to $3k$ entities are randomly sampled from the original graph. 
The two graphs are controlled to share some amounts of common entities. Different levels of entity overlapping are investigated, from $0\%$ (non-overlapping setting) to around $1.5\%, 3\%$, and $5\%$ (overlapping setting).
Moreover, different predicates are removed so that $G^1$ and $G^2$ share the same predicate set, i.e. $\mathcal{R}^1 \equiv \mathcal{R}^2 \equiv \mathcal{R}$.

Intra-domain triplets $(e_i, r_k, e_j)$ whose both entities $e_i, e_j$ belong to the same graph are used for training. Inter-domain triplets $(e_i, r_k, e_j)$ whose entities $e_i, e_j$ belong to different graphs are used for validating and testing inter-domain performance. The validation and test ratio is $20:80$.
Even though the goal is to evaluate a model's ability to perform inter-domain link prediction,
both inter-domain and intra-domain link prediction performances are evaluated. 
This is because the proposed method should improve inter-domain link prediction while does not harm intra-domain link prediction. Therefore, $5\%$ of intra-domain triplets are further spared from the training data for monitoring intra-domain performance.

The details for the case of $3\%$ overlapping are shown in Table~\ref{tab:dataset}. In other cases, the datasets share similar statistics.

\begin{table*}[ht]
\caption{Details of the datasets in the case of $3\%$ overlapping. The other cases share similar statistics.}
\centering
\resizebox{0.85\textwidth}{!}{
    \begin{tabular}{||c||c|c|c|c|c|c|c||}
    \hline
        Datasets & \#Ent G1 & \#Ent G2 & \#Rel & \#Train & \#Inter Valid & \#Intra Test & \#Inter Test  \\
        
    \hline
        FB15k-237 & 2675 & 2677 & 179 & 24.3k & 4.3k & 1.3k & 17.7k\\
    
    \hline
        WN18RR & 2804 & 2720 & 10 & 5.1k & 105 & 148 & 1.1k \\
        
    \hline
        DBbook2014 & 2932 & 2893 & 11 & 34.6k & 6.5k & 1.8k & 26.8k \\
        
    \hline
        ML1M & 2764 & 2726 & 18 & 39.3k & 6.5k & 2k & 27k \\
    
    \hline
    \end{tabular}
}
\label{tab:dataset}
\end{table*}

\subsection{Evaluation methods and Baselines}

In the experiments, Hit@10 score and ROC-AUC score are used for quantifying both inter-domain and intra-domain performances.

\subsubsection{Evaluation with Hit@10.} 
The Hit@10 score is computed by ranking true entities based on their scores. For each true triplet $(e_i, r_k, e_j)$ in the test sets, one entity $e_i$ (or $e_j$) is hidden to create an unfinished triplet $(\cdot, r_k, e_j)$ (or $(e_i, r_k, \cdot)$). All entities $e_\text{cand}$ are used as candidates for completing the unfinished triplet and the scores of $(e_\text{cand}, r_k, e_j)$ (or $(e_i, r_k, e_\text{cand})$) are computed. Note that the candidates $e_\text{cand}$ are taken from the same entity set as $e_i$ (or $e_j$), i.e. if $e_i$ (or $e_j$) $\in \mathcal{E}^t$ then entities $e_\text{cand}$ are taken from $\mathcal{E}^t$.
The ranking of $e_i$ (or $e_j$) is computed according to the scores. The higher ``true" entities are ranked the better a model is at predicting hidden true triplets.
Hit@10 score is used for quantifying the link prediction performance and is calculated as the percentage of ``true" entities being ranked inside the top $10$.

\subsubsection{Evaluation with ROC-AUC.}
In order to compute the ROC-AUC score, triplets in the test set are treated as positive samples. An equal number of triplets are uniformly sampled from the entity sets and the predicate set to create negative samples. Due to the sparsity of each graph, it is safe to consider the sampled triplets as negative. During the sampling process, both sampled entities are controlled to belong to the same graph in the intra-domain case and belong to different graphs in the inter-domain case.

\subsubsection{Evaluated Models.}

In the experiments, RESCAL is used as the baseline method. The proposed method with Wasserstein regularization is denoted as WD while the one with MMD regularization is denoted as MMD.

\subsection{Implementation details}

\subsubsection{Negative sampling.}

Only intra-domain negative triplets are used in order to train the pairwise ranking loss~\eqref{eq:rescal} with SGD, i.e. negative triplet set $\mathcal{D}^-$ only contains negative triplets $(e_l, r_h, e_t)$ whose both entities belong to the same graph.

\subsubsection{Warmstarting.}

Completely learning from scratch might be difficult since the regularizer $M$ can add noise at the early state. Instead, it is beneficial to warmstart the proposed method's embeddings with embeddings roughly learned by RESCAL. Specifically, we run RESCAL for $100$ epochs to learn initial embeddings. After that, to maintain the fairness of equal training time, both the proposed method and RESCAL are warmstarted with the roughly learned embeddings.

\subsubsection{Hyperparameters.}
In the implementation, the latent embedding dimension is set to equal $100$. All experiments are run for $300$ epochs. Early stopping is employed with a patience budget of $50$ epochs. Other hyperparameters, namely $\alpha$, learning rate, and batch size, are tuned on the inter-domain validation set using Optuna~\cite{optuna}. During the tuning process, $\alpha$ is sampled to be between $0.5$ and $10.0$, while the learning rate and batch size are chosen from $\{0.01, 0.005, 0.001, 0.0005\}$ and $\{100, 300, 500, 700\}$, respectively.
The hyperparameters of RESCAL is tuned similarly with fixed $\alpha = 0.0$.
The kernel used in MMD is set to be a mixture of Gaussian kernels with the bandwidth list of $[0.25, 0.5, 1., 2., 4.]*c$ where $c$ is the mean Euclidean distance between the entities.
All results are averaged over $10$ random runs\footnote{The code is available at \url{https://github.com/phucdoitoan/inter-domain_lp}}.

\subsection{Experimental results}

The experimental results are shown in Tables~\ref{tab:inter-hit@10},~\ref{tab:inter-ROC},~\ref{tab:intra-hit@10}, and~\ref{tab:intra-ROC}. Note that a random predictor has a Hit@$10$ score of less than $0.004$ and a ROC-AUC score of around $0.5$.

\subsubsection{Inter-domain link prediction.}

As being demonstrated in tables~\ref{tab:inter-hit@10} and~\ref{tab:inter-ROC}, the proposed method with WD regularizer works well with the FB15k-237 dataset, which outperforms RESCAL in all settings. Especially in the overlapping cases where few entities are shared between the graphs, both Hit@10 and ROC-AUC scores are improved significantly. 
The WD regularizer also demonstrates its usefulness with the DBbook2014 and ML1M datasets. The Hit@10 scores are boosted up in most cases of overlapping settings, while the ROC-AUC scores are consistently enhanced over that of RESCAL. Most of the time, the improvements are considerable. 
However, for the case of the ML1M dataset with $3\%$ overlapping entities, the WD regularizer causes the Hit@10 score to deteriorate, from $0.230$ to $0.213$.
On the other hand, the MMD regularizer seems not to be beneficial for the task. Unexpectedly, the regularizer introduces noise and reduces the accuracy of inter-domain link prediction.
In the case of the WN18RR dataset, both RESCAL and the proposed method fail to perform, in which all Hit@10 and ROC-AUC scores are close to random. This might be due to the extreme sparsity of the dataset, whose amount of observed triplets is only about one-fifth of that of the other datasets.

In all the four datasets, sharing some common entities, even with a small number, is helpful and important for predicting inter-domain links. 
These common entities act as anchors between the graphs, which guide the regularizer to learn similar embedding distributions.
Without common entities, the learning process becomes more challenging and often results in uncertain predictors as being shown in the $0\%$ overlapping cases.
The overlapping setting is reasonable because, in practice, the two graphs often share some amounts of common entities, e.g. the same users and the same popular items reappear in different e-commerce platforms.


\begin{table*}[t]
\caption{\textbf{Inter-domain Hit@$10$} scores.
Italic numbers indicate better results while bold numbers and bold numbers with asterisk $*$ indicate better results at significance level $p=0.1$ and $p=0.05$, respectively.
The proposed method with WD regularizer achieves better scores in many settings.}

\centering
\resizebox{0.8\textwidth}{!}{
    \begin{tabular}{||c|c||>{\centering\arraybackslash}m{2cm}|>{\centering\arraybackslash}m{2cm}|>{\centering\arraybackslash}m{2cm}|>{\centering\arraybackslash}m{2cm}||}
    \hline
            Overlapping & Model & FB15k-237 & WN18RR & DBbook2014 & ML1M \\
    \hline
    \hline
        \multirow{3}{*}{0\%} & RESCAL & 0.110 {\tiny $\pm$0.038} & 0.027 {\tiny $\pm$0.003} & 0.087 {\tiny $\pm$0.058} & 0.062 {\tiny $\pm$0.074}  \\
    \cline{2-6}
                                     & MMD & 0.111 {\tiny $\pm$0.038} & 0.031 {\tiny $\pm$0.004} & 0.085 {\tiny $\pm$0.057} & 0.063 {\tiny $\pm$0.072}  \\
    \cline{2-6}
                                     & WD & \textbf{0.145 {\tiny $\pm$0.063}} & 0.024 {\tiny $\pm$0.004} & 0.084 {\tiny $\pm$0.070} & 0.061 {\tiny $\pm$0.067}  \\
    
    \hline
    \hline
        \multirow{3}{*}{1.5\%} & RESCAL & 0.251 {\tiny $\pm$0.031} & 0.025 {\tiny $\pm$0.002} & 0.107 {\tiny $\pm$0.035} & 0.210 {\tiny $\pm$0.034}  \\
    \cline{2-6}
                                     & MMD & 0.237 {\tiny $\pm$0.043} & 0.026 {\tiny $\pm$0.003} & 0.109 {\tiny $\pm$0.037} & 0.180 {\tiny $\pm$0.067}  \\
    \cline{2-6}
                                     & WD & \textbf{0.291 {\tiny $\pm$0.031}}$^*$ & 0.024 {\tiny $\pm$0.002} & \textit{0.128 {\tiny $\pm$0.059}} & \textbf{0.240 {\tiny $\pm$0.031}}$^*$  \\
    
    \hline                        
    \hline
        \multirow{3}{*}{3\%} & RESCAL & 0.302 {\tiny $\pm$0.020} & 0.028 {\tiny $\pm$0.004} & 0.266 {\tiny $\pm$0.056} & \textbf{0.230 {\tiny $\pm$0.003}}$^*$  \\
    \cline{2-6}
                                     & MMD & 0.292 {\tiny $\pm$0.020} & 0.026 {\tiny $\pm$0.004} & 0.227 {\tiny $\pm$0.081} & 0.228 {\tiny $\pm$0.002}  \\
    \cline{2-6}
                                     & WD & \textbf{0.328 {\tiny $\pm$0.011}}$^*$ & 0.025 {\tiny $\pm$0.004} & \textbf{0.318 {\tiny $\pm$0.066}}$^*$ & 0.213 {\tiny $\pm$0.006}  \\
    
    \hline
    \hline
        \multirow{3}{*}{5\%} & RESCAL & 0.339 {\tiny $\pm$0.007} & 0.027 {\tiny $\pm$0.005} & 0.389 {\tiny $\pm$0.032} & 0.237 {\tiny $\pm$0.011}  \\
    \cline{2-6}
                                     & MMD & 0.334 {\tiny $\pm$0.006} & 0.026 {\tiny $\pm$0.004} & 0.388 {\tiny $\pm$0.027} & 0.236 {\tiny $\pm$0.010}  \\                
    \cline{2-6}
                                     & WD & \textbf{0.361 {\tiny $\pm$0.010}}$^*$ & 0.031 {\tiny $\pm$0.004} & 0.389 {\tiny $\pm$0.051} & \textbf{0.256 {\tiny $\pm$0.006}}$^*$  \\
    
    \hline
    \end{tabular}
}
\label{tab:inter-hit@10}

\end{table*}


\begin{table*}[ht]
\caption{\textbf{Inter-domain  ROC-AUC} scores.
Italic numbers indicate better results while bold numbers and bold numbers with asterisk $*$ indicate better results at significance level $p=0.1$ and $p=0.05$, respectively.
The proposed method with WD regularizer achieves better scores in many settings.}

\centering
\resizebox{0.8\textwidth}{!}{
    \begin{tabular}{||c|c||>{\centering\arraybackslash}m{2cm}|>{\centering\arraybackslash}m{2cm}|>{\centering\arraybackslash}m{2cm}|>{\centering\arraybackslash}m{2cm}||}
    \hline
          Overlapping & Model & FB15k-237 & WN18RR & DBbook2014 & ML1M \\
    \hline
    \hline
        \multirow{3}{*}{0\%} & RESCAL & 0.504 {\tiny $\pm$0.092} & 0.504 {\tiny $\pm$0.009} & 0.483 {\tiny $\pm$0.097} & 0.464 {\tiny $\pm$0.173}  \\
    \cline{2-6}
                                     & MMD & 0.507 {\tiny $\pm$0.093} & 0.500 {\tiny $\pm$0.010} & 0.485 {\tiny $\pm$0.095} & 0.480 {\tiny $\pm$0.172}  \\
    \cline{2-6}
                                     & WD & \textit{0.548 {\tiny $\pm$0.118}} & 0.505 {\tiny $\pm$0.009} & 0.488 {\tiny $\pm$0.099} & 0.495 {\tiny $\pm$0.179}  \\
    
    \hline
    \hline
        \multirow{3}{*}{1.5\%} & RESCAL & 0.793 {\tiny $\pm$0.044} & 0.512 {\tiny $\pm$0.009} & 0.640 {\tiny $\pm$0.066} & 0.805 {\tiny $\pm$0.027}  \\
    \cline{2-6}
                                     & MMD & 0.770 {\tiny $\pm$0.063} & 0.507 {\tiny $\pm$0.009}
 & 0.632 {\tiny $\pm$0.063} & 0.754 {\tiny $\pm$0.087}  \\
    \cline{2-6}
                                     & WD & \textbf{0.837 {\tiny $\pm$0.033}}$^*$ & 0.510 {\tiny $\pm$0.007} & \textit{0.671 {\tiny $\pm$0.087}} & \textbf{0.842 {\tiny $\pm$0.017}}$^*$  \\
    
    \hline                        
    \hline
        \multirow{3}{*}{3\%} & RESCAL & 0.825 {\tiny $\pm$0.022} & 0.503 {\tiny $\pm$0.009} & 0.762 {\tiny $\pm$0.032} & 0.832 {\tiny $\pm$0.006}  \\
    \cline{2-6}
                                     & MMD & 0.813 {\tiny $\pm$0.030} & 0.498 {\tiny $\pm$0.011} & 0.714 {\tiny $\pm$0.060} & 0.831 {\tiny $\pm$0.007}  \\
    \cline{2-6}
                                     & WD & \textbf{0.850 {\tiny $\pm$0.013}}$^*$ & 0.502 {\tiny $\pm$0.013} & \textbf{0.809 {\tiny $\pm$0.030}}$^*$ & 0.840 {\tiny $\pm$0.008}  \\
    
    \hline
    \hline
        \multirow{3}{*}{5\%} & RESCAL & 0.870 {\tiny $\pm$0.008} & 0.498 {\tiny $\pm$0.021} & 0.824 {\tiny $\pm$0.012} & 0.845 {\tiny $\pm$0.007}  \\
    \cline{2-6}
                                     & MMD & 0.875 {\tiny $\pm$0.007} & 0.498 {\tiny $\pm$0.012} & 0.823 {\tiny $\pm$0.015} & 0.845 {\tiny $\pm$0.006}  \\                
    \cline{2-6}
                                     & WD & \textbf{0.902 {\tiny $\pm$0.010}}$^*$ & 0.498 {\tiny $\pm$0.013} & \textbf{0.835 {\tiny $\pm$0.020}} & \textbf{0.867 {\tiny $\pm$0.003}}$^*$  \\
    \hline
    \end{tabular}
}
\label{tab:inter-ROC}

\end{table*}

\subsubsection{Intra-domain link prediction.}

Even though the main goal is to predict inter-domain links, it is preferable that the regularizers do not harm performance on intra-domain link prediction when fusing the two domains' entity distributions.
As being demonstrated in table~\ref{tab:intra-ROC}, the proposed method is able to maintain similar or better intra-domain ROC-AUC scores compared to RESCAL.
However, it sometimes requires trade-offs in terms of the Hit@10 score, which is shown in table~\ref{tab:intra-hit@10}. Specifically, the WD regularizer worsens the intra-domain Hit@10 scores compared to RESCAL in FB15k-237 with $5\%$ overlapping and ML1M with $1.5\%$ overlapping settings despite helping improve the inter-domain counterparts. 
It also hurts the intra-domain Hit@10 score in ML1M with $3\%$ overlapping setting.


\begin{table*}[t]
\caption{\textbf{Intra-domain Hit@$10$} scores.
Bold numbers with asterisk $*$ indicate better results at significance level $p=0.05$.
Generally, the proposed method with WD regularizer preserves the intra-domain Hit@10 scores despite requiring trade-offs in some cases.}

\centering
\resizebox{0.8\textwidth}{!}{
    \begin{tabular}{||c|c||>{\centering\arraybackslash}m{2cm}|>{\centering\arraybackslash}m{2cm}|>{\centering\arraybackslash}m{2cm}|>{\centering\arraybackslash}m{2cm}||}
    \hline
          Overlapping & Model & FB15k-237 & WN18RR & DBbook2014 & ML1M \\
    \hline
    \hline
        \multirow{3}{*}{0\%} & RESCAL & 0.451 {\tiny $\pm$0.031} & 0.418 {\tiny $\pm$0.031} & 0.468 {\tiny $\pm$0.011} & 0.302 {\tiny $\pm$0.076}  \\
    \cline{2-6}
                                     & MMD & 0.461 {\tiny $\pm$0.029} & 0.342 {\tiny $\pm$0.086} & 0.449 {\tiny $\pm$0.012} & 0.307 {\tiny $\pm$0.070}  \\
    \cline{2-6}
                                     & WD & 0.469 {\tiny $\pm$0.019} & 0.421 {\tiny $\pm$0.032} & 0.472 {\tiny $\pm$0.014} & 0.332 {\tiny $\pm$0.027}  \\
    
    \hline
    \hline
        \multirow{3}{*}{1.5\%} & RESCAL & 0.433 {\tiny $\pm$0.008} & 0.390 {\tiny $\pm$0.040} & 0.296 {\tiny $\pm$0.039} & \textbf{0.425 {\tiny $\pm$0.006}}$^*$  \\
    \cline{2-6}
                                     & MMD & 0.438 {\tiny $\pm$0.008} & 0.330 {\tiny $\pm$0.067} & 0.328 {\tiny $\pm$0.027} & 0.423 {\tiny $\pm$0.036}  \\
    \cline{2-6}
                                     & WD & 0.427 {\tiny $\pm$0.009} & 0.408 {\tiny $\pm$0.035} & 0.291 {\tiny $\pm$0.038} & 0.412 {\tiny $\pm$0.008}  \\
    
    \hline                        
    \hline
        \multirow{3}{*}{3\%} & RESCAL & 0.433 {\tiny $\pm$0.009} & 0.476 {\tiny $\pm$0.074} & 0.413 {\tiny $\pm$0.008} & \textbf{0.447 {\tiny $\pm$0.006}}$^*$  \\
    \cline{2-6}
                                     & MMD & 0.447 {\tiny $\pm$0.011} & 0.485 {\tiny $\pm$0.074} & 0.411 {\tiny $\pm$0.017} & 0.444 {\tiny $\pm$0.008}  \\
    \cline{2-6}
                                     & WD & 0.439 {\tiny $\pm$0.009} & \textbf{0.620 {\tiny $\pm$0.026}}$^*$ & 0.412 {\tiny $\pm$0.009} & 0.413 {\tiny $\pm$0.021}  \\
    
    \hline
    \hline
        \multirow{3}{*}{5\%} & RESCAL & \textbf{0.433 {\tiny $\pm$0.009}}$^*$ & 0.455 {\tiny $\pm$0.038} & 0.418 {\tiny $\pm$0.010} & 0.408 {\tiny $\pm$0.005}  \\
    \cline{2-6}
                                     & MMD & 0.421 {\tiny $\pm$0.009} & 0.416 {\tiny $\pm$0.058} & 0.420 {\tiny $\pm$0.014} & 0407 {\tiny $\pm$0.004}  \\                
    \cline{2-6}
                                     & WD & 0.413 {\tiny $\pm$0.007} & 0.479 {\tiny $\pm$0.076} & 0.412 {\tiny $\pm$0.022} & 0.401 {\tiny $\pm$0.005}  \\
    \hline
    \end{tabular}
}
\label{tab:intra-hit@10}

\end{table*}


\begin{table*}[ht]
\caption{\textbf{Intra-domain ROC-AUC} scores. 
Bold numbers with asterisk $*$ indicate better results at significance level $p=0.05$. The propose method maintains similar or better intra-domain ROC-AUC scores compared to RESCAL.}

\centering
\resizebox{0.8\textwidth}{!}{
    \begin{tabular}{||c|c||>{\centering\arraybackslash}m{2cm}|>{\centering\arraybackslash}m{2cm}|>{\centering\arraybackslash}m{2cm}|>{\centering\arraybackslash}m{2cm}||}
    \hline
          Overlapping & Model & FB15k-237 & WN18RR & DBbook2014 & ML1M \\
    \hline
    \hline
        \multirow{3}{*}{0\%} & RESCAL & 0.925 {\tiny $\pm$0.018} & 0.819 {\tiny $\pm$0.018} & 0.915 {\tiny $\pm$0.004} & 0.897 {\tiny $\pm$0.022}  \\
    \cline{2-6}
                                     & MMD & 0.924 {\tiny $\pm$0.018} & 0.818 {\tiny $\pm$0.019} & 0.915 {\tiny $\pm$0.005} & 0.897 {\tiny $\pm$0.035}  \\
    \cline{2-6}
                                     & WD & 0.928 {\tiny $\pm$0.006} & 0.811 {\tiny $\pm$0.017} & 0.918 {\tiny $\pm$0.005} & \textbf{0.932 {\tiny $\pm$0.004}}$^*$  \\
    
    \hline
    \hline
        \multirow{3}{*}{1.5\%} & RESCAL & 0.929 {\tiny $\pm$0.003} & 0.814 {\tiny $\pm$0.018} & 0.871 {\tiny $\pm$0.032} & 0.950 {\tiny $\pm$0.003}  \\
    \cline{2-6}
                                     & MMD & 0.931 {\tiny $\pm$0.003} & 0.807 {\tiny $\pm$0.029} & 0.892 {\tiny $\pm$0.009} & 0.954 {\tiny $\pm$0.003}  \\
    \cline{2-6}
                                     & WD & 0.932 {\tiny $\pm$0.006} & 0.818 {\tiny $\pm$0.020} & 0.868 {\tiny $\pm$0.040} & 0.954 {\tiny $\pm$0.002}  \\
    
    \hline                        
    \hline
        \multirow{3}{*}{3\%} & RESCAL & 0.922 {\tiny $\pm$0.006} & 0.870 {\tiny $\pm$0.018} & 0.885 {\tiny $\pm$0.008} & 0.946 {\tiny $\pm$0.005}  \\
    \cline{2-6}
                                     & MMD & 0.926 {\tiny $\pm$0.005} & 0.861 {\tiny $\pm$0.011} & 0.877 {\tiny $\pm$0.026} & 0.948 {\tiny $\pm$0.003}  \\
    \cline{2-6}
                                     & WD & 0.921 {\tiny $\pm$0.007} & 0.860 {\tiny $\pm$0.018} & 0.890 {\tiny $\pm$0.005} & 0.949 {\tiny $\pm$0.003}  \\
    
    \hline
    \hline
        \multirow{3}{*}{5\%} & RESCAL & 0.927 {\tiny $\pm$0.007} & 0.869 {\tiny $\pm$0.007} & 0.878 {\tiny $\pm$0.008} & 0.949 {\tiny $\pm$0.003}  \\
    \cline{2-6}
                                     & MMD & 0.935 {\tiny $\pm$0.005} & 0.835 {\tiny $\pm$0.050} & 0.879 {\tiny $\pm$0.008} & 0.952 {\tiny $\pm$0.003}  \\                
    \cline{2-6}
                                     & WD & \textbf{0.937 {\tiny $\pm$0.004}}$^*$ & 0.860 {\tiny $\pm$0.020} & \textbf{0.885 {\tiny $\pm$0.009}}$^*$ & 0.953 {\tiny $\pm$0.003}  \\
    \hline
    \end{tabular}
}
\label{tab:intra-ROC}

\end{table*}

\subsubsection{Summary.}

The proposed method with WD regularizer significantly improves the performance of inter-domain link prediction over the baseline method while being able to preserve the intra-domain performance in the FB15k-237 and DBbook2014 datasets. In the ML1M dataset, it benefits the inter-domain performance at the risk of decreasing intra-domain Hit@10 scores.
Unexpectedly, the MMD regularizer does not work well and empirically causes deterioration of the inter-domain performance.
These negative results might be due to local optimal arising when minimizing MMD with a finite number of samples, as recently studied in~\cite{coulomb_AE}. Further detailed analysis would be necessary before one can firmly judge the performance of the MMD regularizer. We leave this matter for future works.
It is also worth mentioning that, in the experiment setting, the sampling of $G^1$ and $G^2$ is repeated independently for each overlapping level.
Therefore, it is not necessary for the link prediction scores to monotonically increase when the overlapping level increases.

\subsubsection{Embedding visualization.}

\begin{figure}[t]
\centering
\subfloat[][Learned with RESCAL]{
  \centering
  \includegraphics[width=0.5\linewidth]{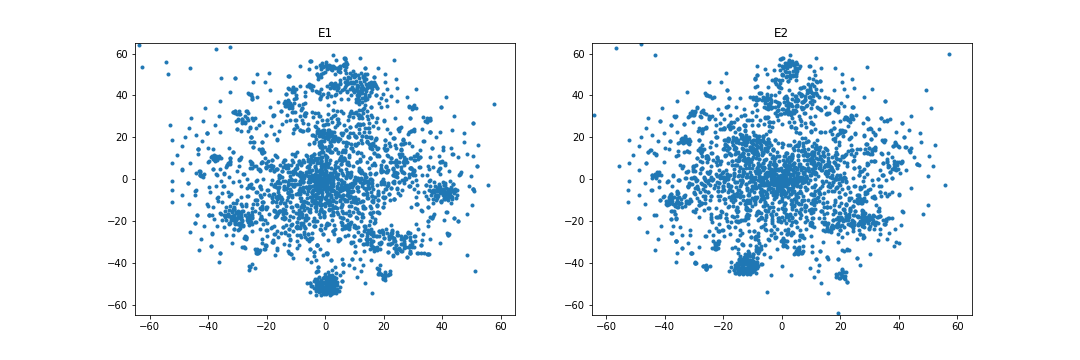}  
  \label{fig:fb15k_emb_rescal}
}
\subfloat[][Learned with RESCAL]{
  \includegraphics[width=0.5\linewidth]{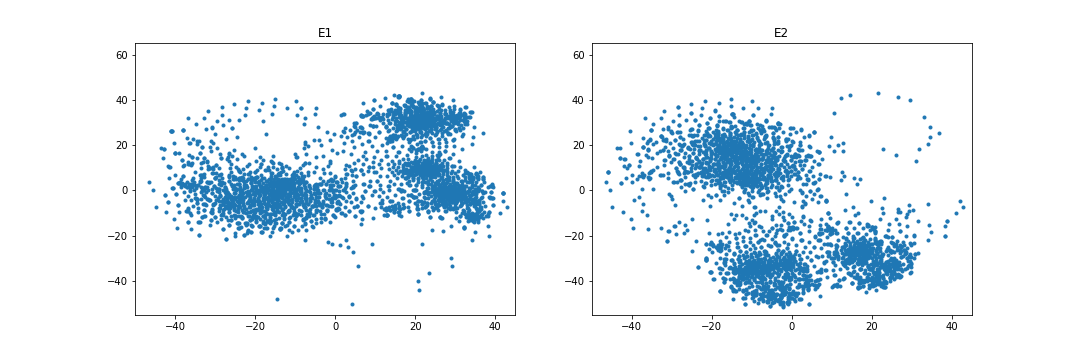}  
  \label{fig:dbbook_emb_rescal}
}

\subfloat[][Learned with WD regularizer]{
  \includegraphics[width=0.5\linewidth]{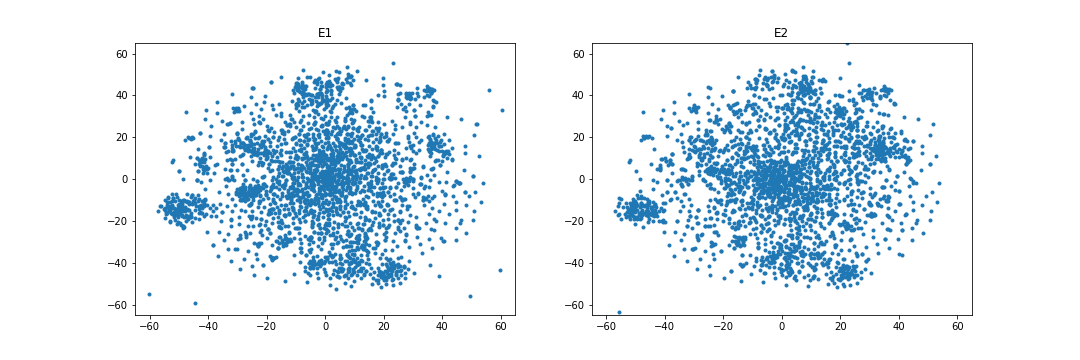}  
  \label{fig:fb15k_emb_ot}
}
\subfloat[][Learned with WD regularizer]{
  \includegraphics[width=0.5\linewidth]{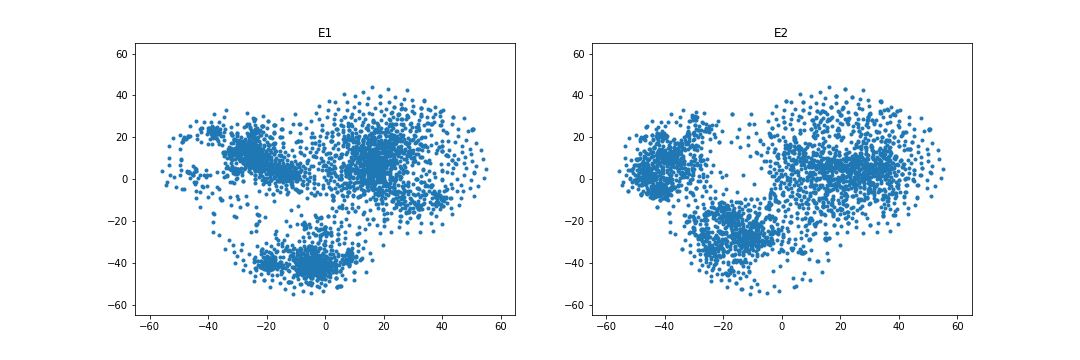}  
  \label{fig:dbbook_emb_ot}
}
\caption{Embedding visualization of FB15k-237 (subfigures a and c) and DBbook2014 (subfigures b and d) datasets with $3\%$ overlapping. The proposed method learns more identical embedding distributions across both domains.}
\label{fig:emb_fb15k_dbbook}
\end{figure}

\begin{figure}[t]
\centering
\subfloat[][Learned with RESCAL]{
  \includegraphics[width=0.5\linewidth]{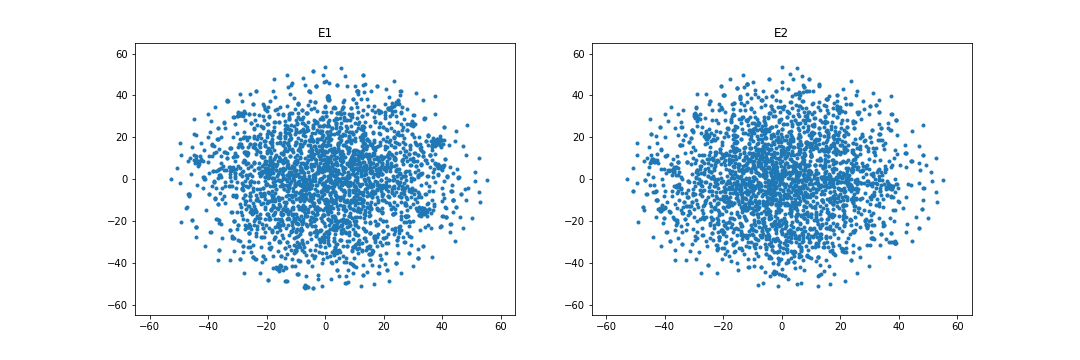}  
  \label{fig:wn18rr_emb_rescal}
}
\subfloat[][Learned with RESCAL]{
  \includegraphics[width=0.5\linewidth]{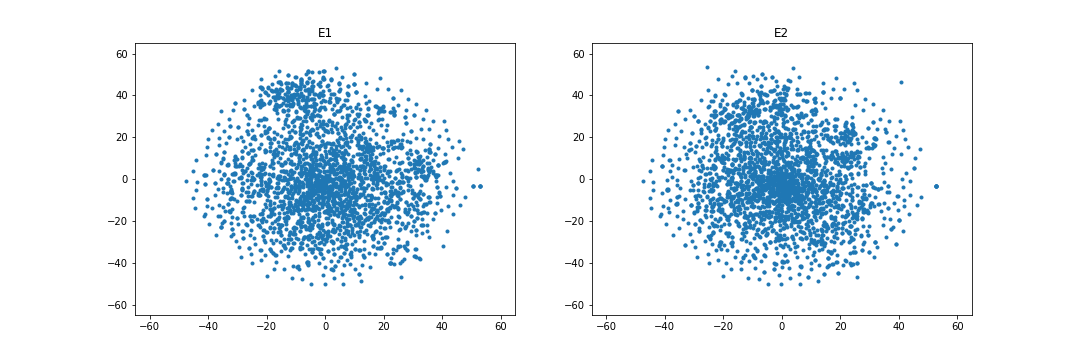}  
  \label{fig:ml1m_emb_rescal}
}

\subfloat[][Learned with WD regularizer]{
  \includegraphics[width=0.5\linewidth]{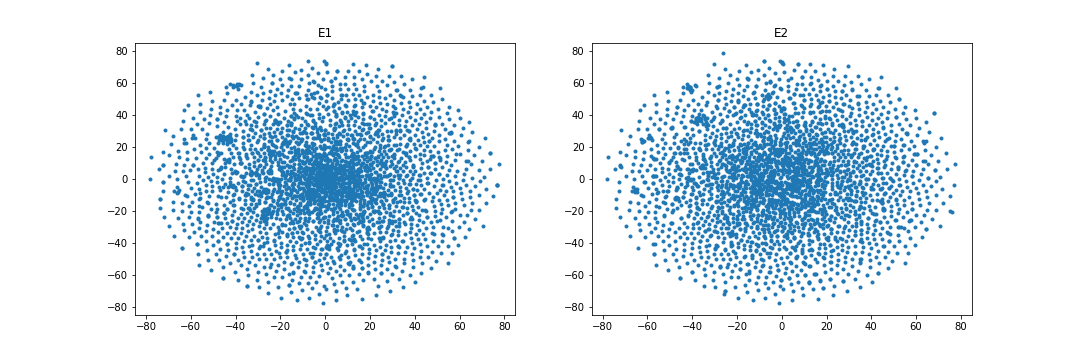}  
  \label{fig:wn18rr__emb_ot}
}
\subfloat[][Learned with WD regularizer]{
  \includegraphics[width=0.5\linewidth]{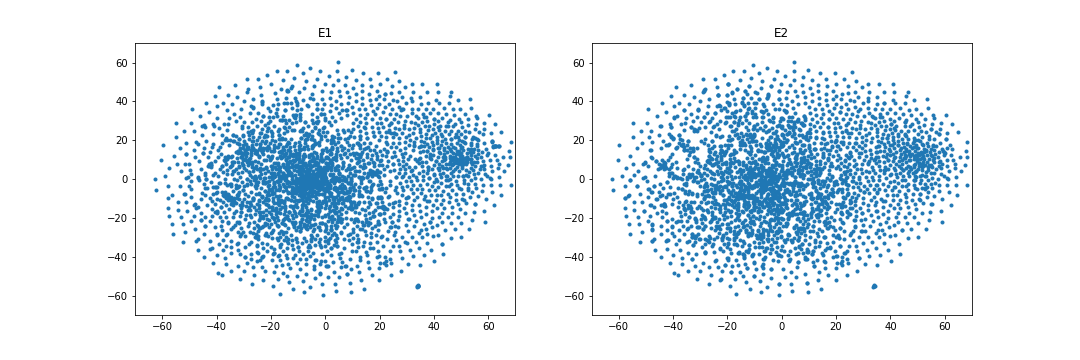}  
  \label{fig:ml1m_emb_ot}
}
\caption{Embedding visualization of WN18RR (subfigures a and c) and ML1M (subfigures b and d) datasets with $3\%$ overlapping. RESCAL is able to learn similar embedding distributions between the two domains while the proposed method seems to add more noise.}
\label{fig:emb_wn18rr_ml1m}
\end{figure}

Figures \ref{fig:emb_fb15k_dbbook} and \ref{fig:emb_wn18rr_ml1m} visualize the entity embeddings learned by RESCAL and the WD regularizer in the case of $3\%$ overlapping. As being seen in Figure \ref{fig:emb_fb15k_dbbook}, WD can learn more identical embedding distributions than RESCAL in the case of the FB15k-237 and DBbook2014 datasets. Especially, in the DBbook2014 dataset, RESCAL can only learn similar shape distributions, but the regularizer can learn distributions with both similar shape and close absolute position. 
However, as being shown in Figure \ref{fig:emb_wn18rr_ml1m}, in the WN18RR and ML1M datasets, the WD regularizer seems to only add noise when learning the embeddings, which results in no improvement or even degradation of both intra-domain and inter-domain Hit@10 scores.

\section{Related Work}

In recent years, the embedding-based approach has become popular in dealing with the link prediction task on a multi-relational knowledge graph (intra-domain). One of the pioneering works in this direction is TransE~\cite{transE}. It is a translation model whose each predicate type corresponds to a translation between the entities' embedding vectors. The model is suitable for $1$-to-$1$ relationships only. Following models such as TransH, TransR, and TransD~\cite{transH,transR,transD} are designed to deal with $n$-to-$1$, $1$-to-$n$, and $n$-to-$n$ relationships. 
Furthermore, tensor-based models such that RESCAL, DistMult, and SimplE~\cite{rescal,disMult,simplE} also gain huge interest. They interpret multi-relational knowledge graphs as $3$-D tensors and employ tensor factorization to learn the entity and predicate embeddings. 
Besides, neural network and complex vector-based models~\cite{NTN,simplEx} are also introduced in the literature. Further details can be found in~\cite{embedding_model_review}.

To the best of our knowledge, the proposed method is the first to consider the inter-domain link prediction problem between multi-relational graphs.
Existing methods in the literature do not directly deal with the problem. The closest line of research focuses on entity alignment in multilingual knowledge graphs, which often aims to match words of the same meanings between different languages. The first work in this line of research is MTransE~\cite{MtransE}. It employs TransE to independently embed different knowledge graphs and perform matching on the embedding spaces. Other methods like JAPE~\cite{JAPE} and BootEA~\cite{BootEA} further improve MTransE by exploiting additional attributes or description information and bootstrapping strategy. MRAEA~\cite{MRAEA} directly learns multilingual entity embeddings by attending over the entities' neighbors and their meta semantic information. Other methods~\cite{MuGNN,deep_graph_consensus} apply Graph Neural Networks for learning alignment-oriented embeddings and achieve state-of-the-art results in many datasets.
All these entity-matching methods implicitly assume most entities in one graph to have corresponding counterparts in the other graph, e.g. words in one lingual graph to have the same meaning words in the other lingual graph. Meanwhile, the proposed method only assumes the similarity between entity distributions.

Minimizing a dissimilarity criterion between distributions is a popular strategy for distribution matching and entity alignment problems. Cao et al. propose Distribution Matching Machines~\cite{DMM} that optimizes maximum mean discrepancy (MMD) between source and target domains for unsupervised domain adaptation tasks. The criterion is successfully applied in distribution matching and domain confusion tasks as well~\cite{MMD_distribution_matching,MMD_domain_confusion}. Besides Wasserstein distance (WD), Gromov-Wasserstein distance (GWD)~\cite{gromov-wasserstein} also is a popular optimal transport metric. It measures the topological dissimilarity between distributions lying on different domains. GWD often requires much heavier computation than WD due to nested loops of Sinkhorn algorithm in current implementations~\cite{gromov-wasserstein}.
Applying optimal transport into the graph matching problem, Xu et al. propose Gromov-Wasserstein Learning framework~\cite{gw-learning} for learning node embedding and node alignment simultaneously, and achieve state of the art in various graph matching datasets. Chen et al.~\cite{got} propose Graph Optimal Transport framework that combines both WD and GWD for entity alignment. The framework is shown to be effective in many tasks such as image-text retrieval, visual question answering, text generation, and machine translation. Due to the computational complexity of GWD, each domain considered in~\cite{gw-learning,got} only contains less than several hundred entities. 
Phuc et al.~\cite{self-cite} propose to apply WD to solve the link prediction problem on two graphs simultaneously. In terms of technical idea, the method is the most similar to the proposed method; however, it only focuses on the intra-domain link prediction problem on undirected homogeneous graphs and requires most of the nodes in one graph to have corresponding counterparts in the other graph.

\section{Conclusion and Future Work}

Inter-domain link prediction is an important task for constructing large multi-relational graphs from smaller related ones. However, existing methods in the literature do not directly address this problem. In this paper, we propose a new approach for the problem via jointly minimizing a divergence between entity distributions during the embedding learning process. Two regularizers have been investigated, in which the WD-based regularizer shows promising results and improves inter-domain link prediction performance considerably.
For future works, we would like to verify the proposed method's effectiveness using more baseline embedding methods besides RESCAL. 
Further analysis on the performance of the MMD-based regularizer will also be conducted.
Moreover, the proposed method currently assumes that both domains share the same underlying entity distribution. This assumption is violated when the domains' entity distributions are not completely identical but partially different. One possible direction for further research is to adopt unbalanced optimal transport as the regularizer, which flexibly allows mass destruction and mass creation between distributions.

\section*{Acknowledgements}

The authors would like to thank the anonymous reviewers for their insightful suggestions and constructive feedback.

%
%
%
%


\end{document}